\documentclass{bmvc2k}
\usepackage[noend]{algorithm2e}

\title{Beef Cattle Instance Segmentation Using Fully Convolutional Neural Network}

\addauthor{Aram Ter-Sarkisov}{aram.ter-sarkisov@dit.ie}{1}
\addauthor{John Kelleher}{john.kelleher@dit.ie}{1}
\addauthor{Bernadette Earley}{bernadette.earley@teagasc.ie}{2}
\addauthor{Michael Keane}{michael.keane@teagasc.ie}{2}
\addauthor{Robert Ross}{robert.ross@dit.ie}{1}
\addinstitution{
 School of Computing \\
 Dublin Institute of Technology\\
 Dublin, Ireland
}
\addinstitution{
 TEAGASC\\
 Grange, Dunsany, Co. Meath, Ireland\\
}
\runninghead{Ter-Sarkisov et al.}{Beef Cattle Instance Segmentation Using FCN}

\begin{document}
\maketitle
\begin{abstract}
In this paper we present a novel instance segmentation algorithm that extends a fully convolutional network to learn to label objects separately without prediction of regions of interest.  We trained the new algorithm on a challenging CCTV recording of beef cattle, as well as benchmark MS COCO and Pascal VOC datasets. Extensive experimentation showed that our approach outperforms the state-of-the-art solutions by up to 8\% on our data. 
\end{abstract}
\section{Introduction}
Recently deep convolutional neural networks have made strong contribution in a number of areas in computer vision. In particular, three areas  have seen progress: object detection, class (semantic) segmentation and instance segmentation (mask representation). Object detection algorithms, such as \cite{girshick2014rich,ren2015faster, girshick2014fast}, generate bounding box proposals for \textit{every} object in the image and  classify these proposals. Class (semantic) segmentation algorithms delineate \textit{classes} of objects at a pixelwise level without making a distinction between two objects belonging to the same class \cite{long2015fully, zheng2015conditional}. Instance segmentation algorithms find a \textit{mask representation} for every object in the image. Most mask segmentation algorithms, such as \cite{he2017mask, dai2016instance, li2016fully}, use a three-stage approach: (a) identify promising regions (bounding box around the object) using a region proposal network (RPN), (b) predict the object class, and (c) extract its mask (often using a fully convolutional network). Recent results on benchmark datasets show the viability of this approach, and currently it is considered to be the state-of-the-art method for extraction of mask representations. The well-regarded Mask R-CNN model uses an RPN to predict bounding box coordinates for the objects in the image and their score, then refines these predictions using a Region of Interest Align (RoI Align) tool, extracts a mask and predicts the class of the object, \cite{he2017mask}. Currently, Mask R-CNN is the publicly available model with the highest reported overall mean average precision on MS COCO 2017 dataset.\\
\linebreak    
Another approach to instance segmentation that has seen growing interest in the past years is instance embedding, which does not make use of bounding boxes or regional prediction at all. Instead, discriminative loss function, like the ones in \cite{de2017semantic, DBLP:journals/corr/abs-1712-08273}  uses feature clustering to locate the centers of instances.\\
\linebreak
We introduce a new framework that uses the output of the fully convolutional network, \cite{long2015fully} and the ground truth mask to refine the mask representation of animals in images without bounding box prediction. Our approach is conceptually simple and does not make use of bounding boxes or RPNs, yet manages to achieve results that compare favourably with all state-of-the-art networks. We show the strength of our approach on Pascal VOC, MS COCO datasets and on a specialized dataset we built from a raw video. The framework, MaskSplitter, learns to output three different types of mask representations: two \textit{bad} and one \textit{good}. A \textit{good} mask representation the one which overlaps with exactly one animal, a \textit{bad} mask representation of the first type overlaps with two or more animals, and a \textit{bad} mask representation of the second type occurs when multiple predicted masks overlap with a single animal. The framework consists of the algorithm that determines the type of the mask representations and the number of true objects (cows) that they predict; pixelwise sigmoid and Euclidean loss functions; and a set of convolutional and fully connected layers, one for every type of prediction. The overhead of the network on top of FCN8s is very small, $\sim$187000 parameters.\\ 
\linebreak
This work is a step towards real-time animal monitoring in farming environments that have different applications, such as early lameness detection, animal welfare improvement and reduction of cattle maintenance costs. For examples of other recent applications of deep learning in animal husbandry and agriculture see, e.g. \cite{stern2015analyzing, gomez2016animal}. The rest of the article is structured in the following way: Section \ref{sec:data} introduces the dataset we created from the video feed at the cattle facility, as well as MS COCO and PASCAL VOC datasets, Section \ref{sec:networks} introduces the MaskSplitter algorithm and the network architecture in which it is embedded, Section \ref{sec:results} presents all experimental results, for FCN + MaskSplitter, as well as Mask R-CNN, section \ref{sec:final} concludes.
\section{Datasets}
\label{sec:data}
\begin{figure}
\begin{center}
\fbox{\includegraphics[width=0.9\linewidth]{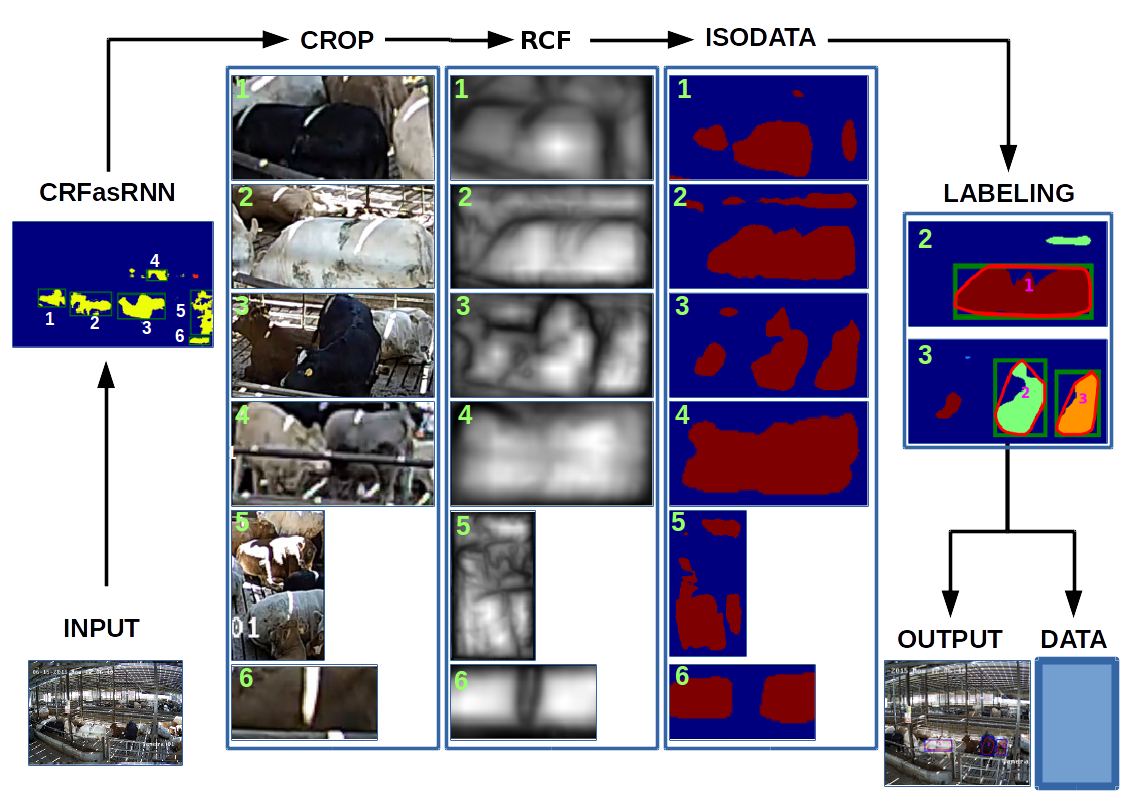}}
\end{center}
\caption{Flowchart of the dataset construction and naive segmentation pipeline. RCF is Richer Convolutional Features CNN from \cite{liu2017richer}, ISODATA is a threshold function from \cite{ridler1978picture}}
\label{fig:ds_construction}
\end{figure}

\begin{figure*}[t]
\centering
\begin{tabular}{cccc}
\bmvaHangBox{\fbox{\includegraphics[width=0.2\linewidth]{img_000000526706}}}&
\bmvaHangBox{\fbox{\includegraphics[width=0.2\linewidth]{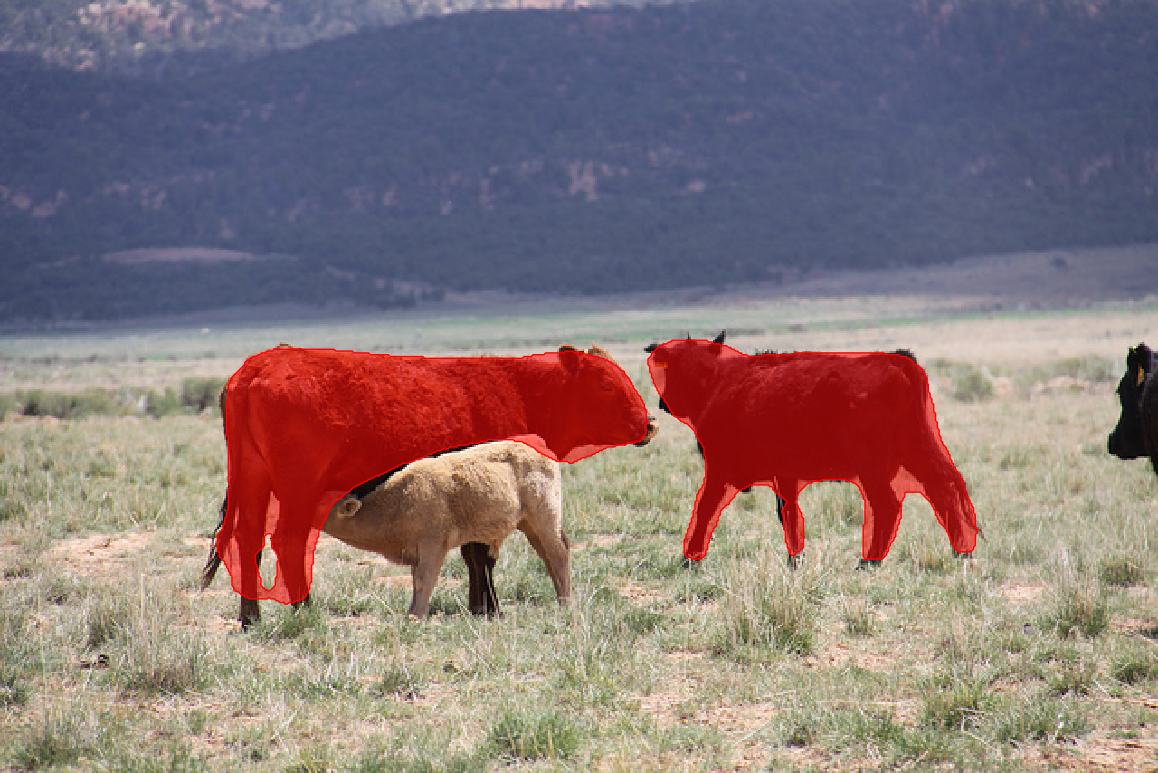}}}&
\bmvaHangBox{\fbox{\includegraphics[width=0.2\linewidth]{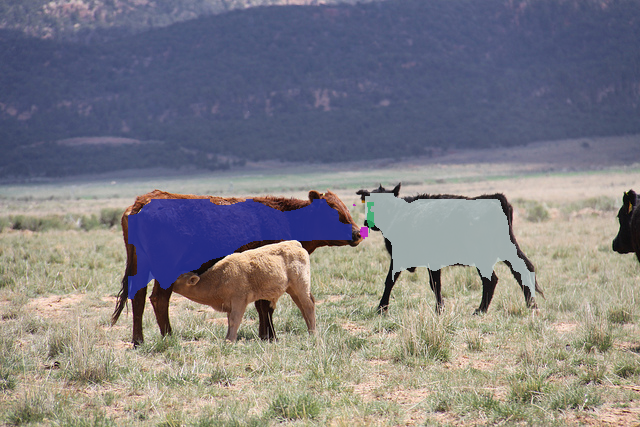}}}&
\bmvaHangBox{\fbox{\includegraphics[width=0.2\linewidth]{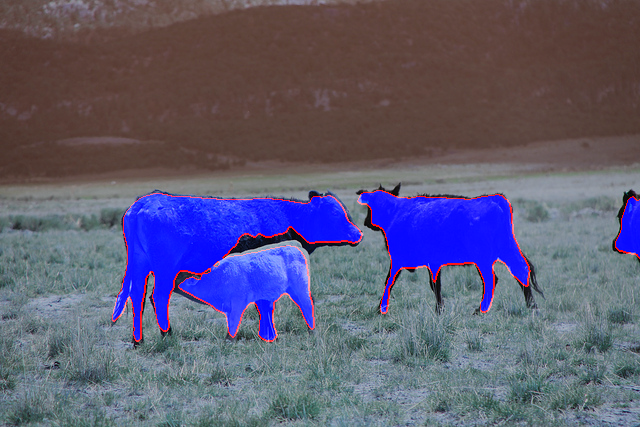}}}\\
\bmvaHangBox{\fbox{\includegraphics[width=0.2\linewidth]{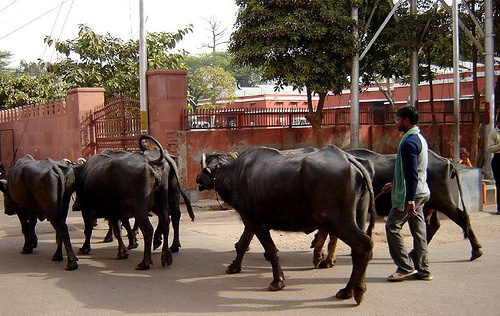}}}&
\bmvaHangBox{\fbox{\includegraphics[width=0.2\linewidth]{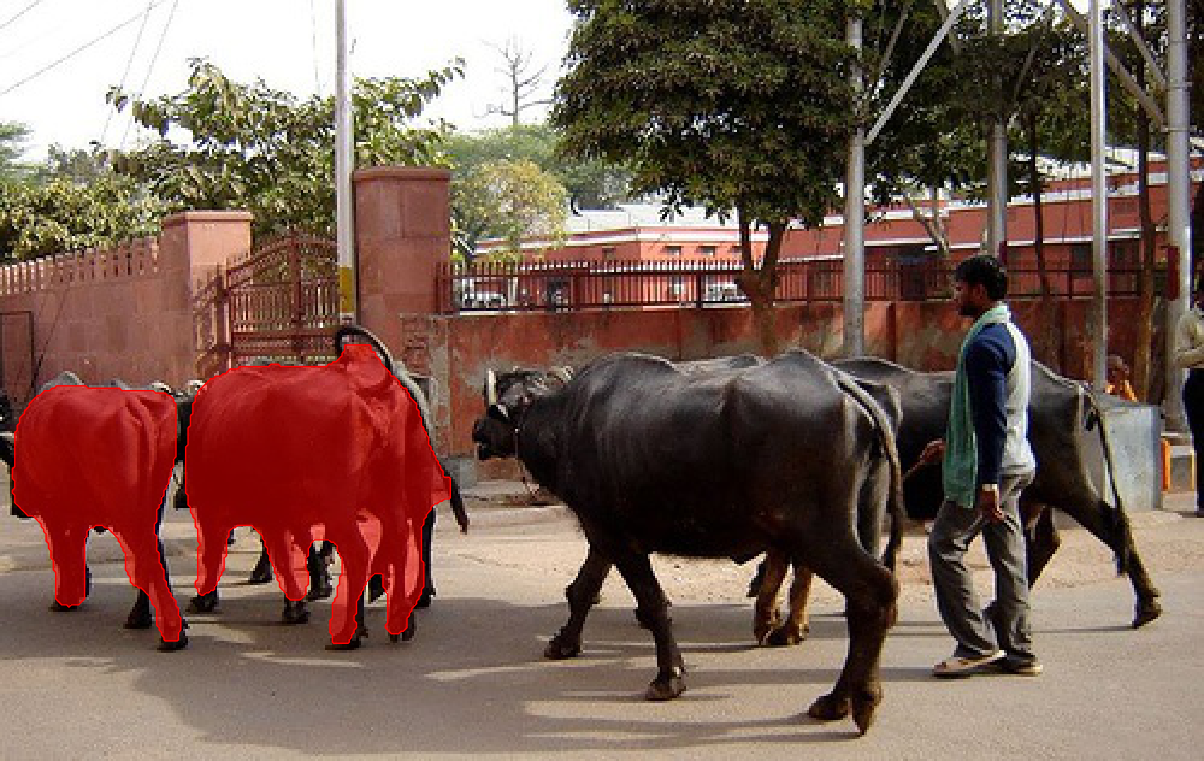}}}&
\bmvaHangBox{\fbox{\includegraphics[width=0.2\linewidth]{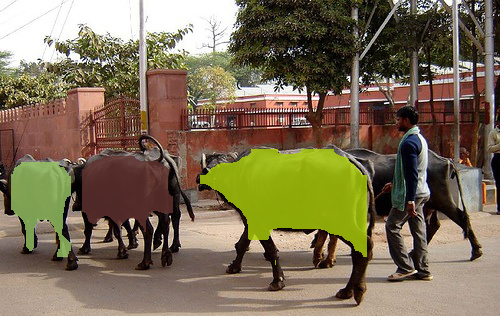}}}&
\bmvaHangBox{\fbox{\includegraphics[width=0.2\linewidth]{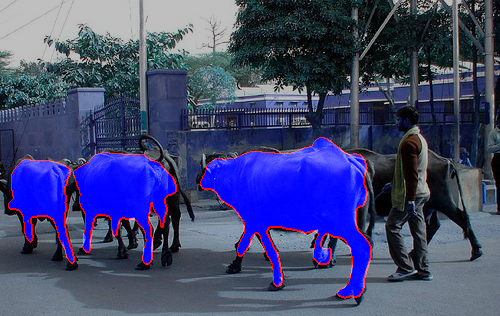}}}\\
Image&Mask R-CNN&FCIS&FCN8s+MS\\
\linebreak
\end{tabular}
\caption{Illustration of Mask R-CNN, FCIS, FCN8s+MaskSplitter performance on MS COCO and Pascal VOC validation datasets}
\label{fig:res_coco_pascal}
\end{figure*}

\begin{figure*}[t]
\centering
\begin{tabular}{cccc}
\bmvaHangBox{\fbox{\includegraphics[width=0.2\linewidth]{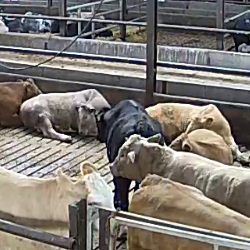}}}&
\bmvaHangBox{\fbox{\includegraphics[width=0.2\linewidth]{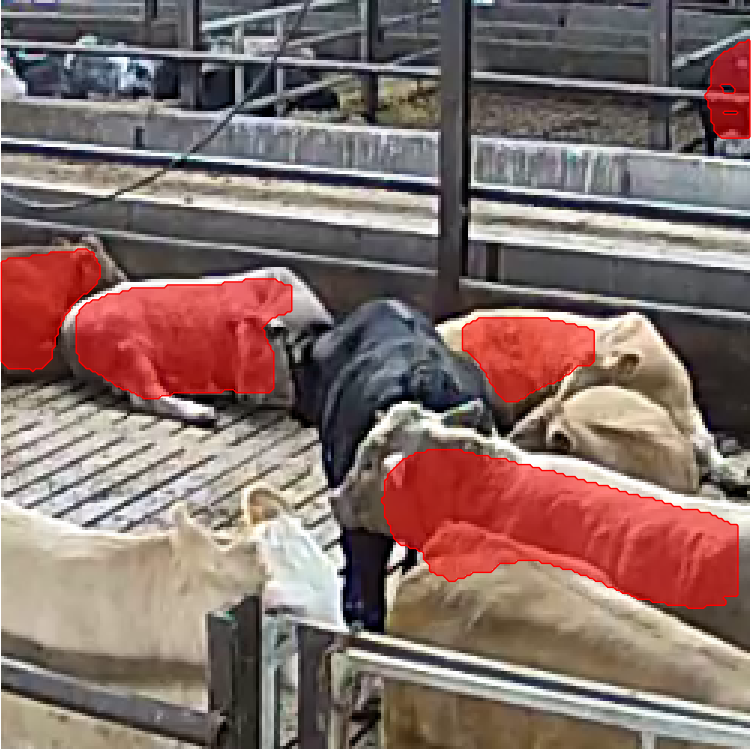}}}&
\bmvaHangBox{\fbox{\includegraphics[width=0.2\linewidth]{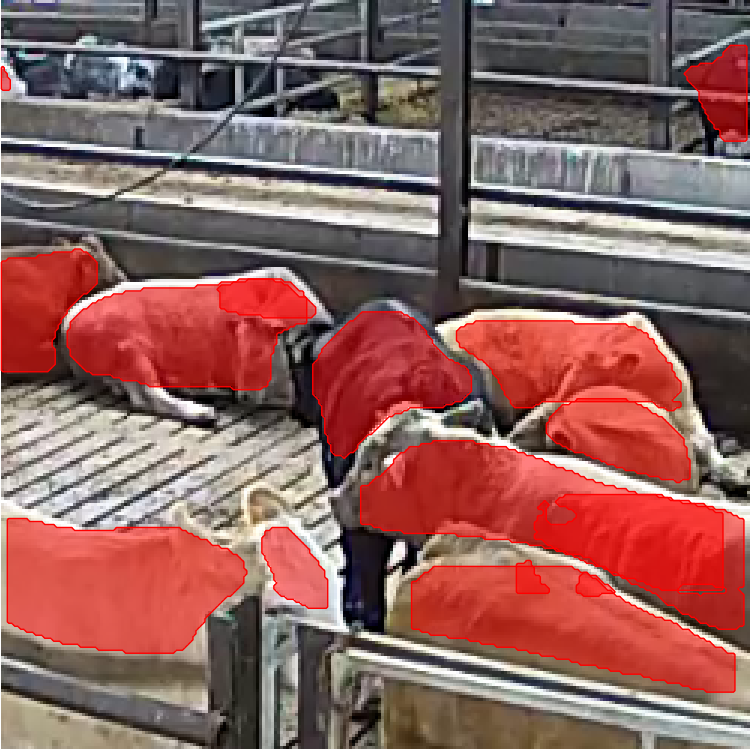}}}&
\bmvaHangBox{\fbox{\includegraphics[width=0.2\linewidth]{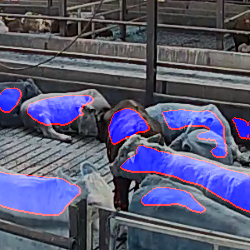}}}\\
Image&Mask R-CNN&Mask R-CNN (heads)&FCN8s+MS\\
\linebreak
\end{tabular}
\caption{Illustration of Mask R-CNN, Mask R-CNN (heads only) and FCN8s+MaskSplitter performance on cattle facility CCTV test dataset (after finetuning)}
\label{fig:res_ours}
\end{figure*}

Pascal VOC 2012 (\cite{everingham2010pascal}) and MS COCO 2017 (\cite{lin2014microsoft}) are two benchmark datasets that are widely used in the Computer Visions community. As we are developing a cattle segmenter, we are interested in cow images in these datasets. There is a total of 1986 images with cows in the MS COCO training set and 87 in the validation set. In Pascal VOC training set there are 64 images with cows and in the validation set there are 71 images. In the training stage we only used MS COCO images. Testing was done both on Pascal VOC and MS COCO datasets.\\       
The CCTV raw videos that were recorded at a winter finishing feedlot for beef heifers over a two-week period on video cameras pointing at a fixed angle towards enclosures with 10 heifers each, a total of 24 pens were included in the study (\cite{keane2017effect}). One of the cattle pens was selected for dataset construction. This video sequence is challenging for a number of reasons, each of which poses a particular challenge to segmenter:
\begin{enumerate}
\item \textbf{The objects} The animals change position frequently and assume different postures when standing up and lying down. Unlike static objects, e.g. TV sets or buildings, their shape changes significantly when an animal moves around, lies down, eats, walks and grooms itself or other pen mates. Hence the network needs to have a much higher generalization capacity.
\item \textbf{Similarity between objects}: it is often difficult (or in fact impossible) to distinguish between two particular animals due to the same coat, color and absense of other distinct physical markings. In the case of partial occlusion, it is nearly impossible even for a human eye to tell one from another.    
\item  \textbf{Occlusion}: animals were allotted to 24 individual pens, each containing 10 heifers, located in the same housing facility. CCTV cameras were positioned over each pen. The pens were evenly distributed in three separate rows, throughout the house. To achieve the respective space allowances, the dimensions of the pens were 4.50 m $\times$ 6.66 m (30 m2) for treatment 1, 9.0 m $\times$ 5.0 m (45 m2) for treatment 2 and 9.0 m $\times$ 6.66 m (60 m2) for treatments 3 and 4. The feed face length was the same (4.5 m) for all pens. As a result of this setup, the view of animals from the cameras suffers from a great degree of partial occlusion.    
\item \textbf{Lighting}: The house had both artificial and natural lighting through vents positioned in the overhead roof. Overall it is rather dark, but there are gaps in the ceiling which allow light through; this is a standard feature of animal housing design. This type of lighting poses a challenge to machine learning algorithms, as they can erroneously learn that these patches or shadows are animals' features.
\item \textbf{Background}: the background (enclosure boundaries) in the video is dark and/or noisy and often cannot be distinguished from the segmented object, because heifers have a color pattern (dark/light brown or grey/white) similar to the background. In many cases the segmented animal instance nearly blends with the background making it a challenge to be detected even by the most advanced algorithms.  
\end{enumerate} 
Considering the substantial challenges, in \cite{ter2017bootstrapping} we developed a separate naive segmentation algorithm that extracts images and ground truth from frames in the video. The algorithm is presented as a flowchart in  Figure \ref{fig:ds_construction}.   
The dataset constructed from this raw video data consists of about 500 frames extracted from one of the feedlots, each frame was cropped to size 250x250 around every animal in the frame and split randomly into the training and validation datasets, of size 4872 and 984. To show that this data is very unique, we tested the state-of-the art networks (Table \ref{tab:results_cowdata_noft}) on the validation set. Results for all thresholds of IoU are far lower than those for the benchmark sets. The selected size of the crops, 250x250 was solely due to the restrictions of VRAM for training of the models. Deployment/testing of the models can be done with image input of any size, and is very fast, including the results in Figures \ref{fig:res_coco_pascal} and \ref{fig:res_ours}.     
\section{Networks}
\label{sec:networks}
\begin{figure}
\centering
\fbox{\includegraphics[width=1\linewidth]{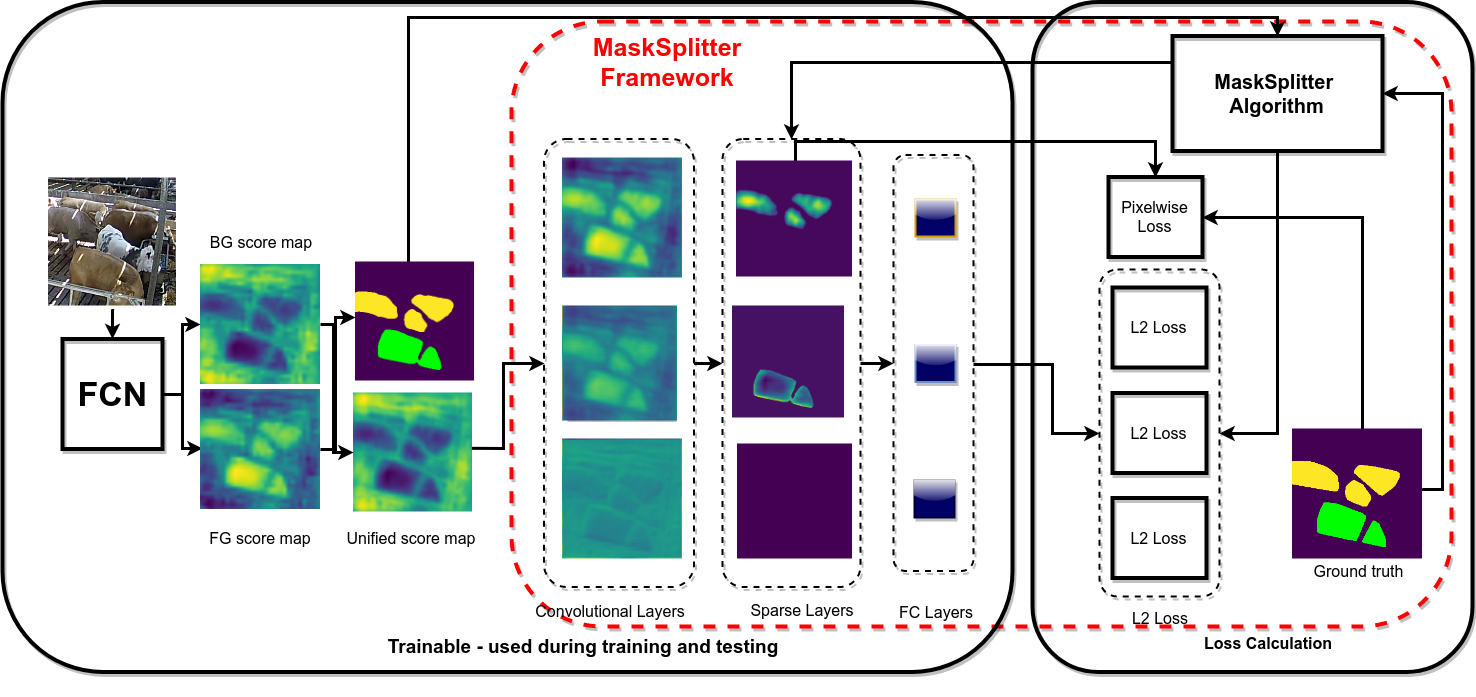}}
\caption{MaskSplitter framework architecture. FCN denotes the Fully Convolutional Network, L2 is Euclidean loss layer. The `bad' prediction for two animals in the binary mask is marked green, as well as the corresponding two animals in the ground truth mask. Mask Splitter has six outputs in total: for each type of prediction it outputs one full-image mask with predictions of this types, extracting it from the binary mask and one number, extracting it from the ground truth map based on the overlaps between predictions and true objects. Best viewed in color.}
\label{fig:masksplitter}
\end{figure}

Although deeper networks like ResNet101, introduced in \cite{he2016deep}, have the capacity to learn more features, FCN, based on VGG16 \cite{simonyan2014very}, has a very convenient architecture that upsamples semantic features to the size of the input image, in effect producing image-sized \textit{score maps}, adjusted for the number of classes. Therefore in all of our experiments we use FCN8s \cite{long2015fully}, the most refined of FCN architectures (others being FCN-16s and FCN-32s) as a backbone network (feature and score map extractor). The output of FCN8s consists of two \textit{score maps} (object and background score maps) that have the same dimensionality as the input image and accumulate pixelwise information about the corresponding class (i.e. either animal or background). If we take a pixel-wise  \textbf{argmax} value of the score map layer, this results in a binary output \textit{mask} with pixels valued at either 0 (background) or 1 (object), depending on the greater score. Each isolated group of pixels in the mask equal to 1 constitutes a \textit{prediction} (\textit{mask representation}) of the animal's instance. This binary mask is the main output of the FCN. Our framework is a refinement of this predictive mask that attempts to recover the representation of every animal.\\
\linebreak
The main concept of MaskSplitter is to separate the learning of three types of predictions: one \textit{good} prediction, and two \textit{bad} predictions. All representations are taken from the FCN8s final binarized score map. Good masks overlaps with exactly one animal in the ground truth map. In case a mask overlaps with two or more cows or two or more masks overlap with one cow, these are identified as bad masks, resp. Type 1 and Type 2. All other masks (i.e. without any overlaps) are ignored. The first step of MaskSplitter is the computation of the IoU matrix with the number of rows equal to the number of predicted masks and the number of columns to the number of true masks/cows. If there is a unique overlap of a predicted mask with a cow, this predicted mask is added to the list of good masks. If this overlap is not unique, all predicted masks are added to the list of bad masks Type 2. If a predicted mask overlaps with two or more cows, it is added to the list of bad masks Type 1. As a result, the algorithm has 6 outputs: 3 values (number of good predictions, number of bad predictions Type 1, number of bad predictions Type 2) and three groups of masks of each type. Thus MaskSplitter splits the original binarized map into three classes.
\label{sec:intro}
\RestyleAlgo{boxruled}
\begin{algorithm}[h]    
   {\bfseries Inputs:}\\
   FCN8s output binary mask $\mathbf{B}$, Ground truth mask $\mathbf{C}$, IoU matrix for all predictions in $\mathbf{B}$ and cows in $\mathbf{C}$\\    
   Number of good predictions = 0, Number of bad predictions (Type 1) = 0, Number of bad predictions (Type 2) = 0\\ 
   $L_1: $Empty list of seen animals (Type 1), $L_2: $ Empty list of seen animals(Type 2)\\
   \For{each prediction $\mathbf{b} \in \mathbf{B}$}   
         {Rank IoUs of this prediction in the decreasing order\\
         $\mathbf{c^{\ast}: argmax{ \ \mathbf{IoU}(\mathbf{b,:})}}$\\
         \If{ this prediction overlaps with only one animal} 	    
         	{\For{each ground truth blob $\mathbf{c} \in  \mathbf{C}$}
	           {Get IoU($\mathbf{b,c}$)\\       
    	       \If{$|\mathbf{IoU}(\mathbf{:,c^{\ast}})|==1$}
                	{Number of good predictions +=1\\
                 	Copy the prediction to the mask with good predictions \\
             	    \textbf{break}}
	           \ElseIf{$|\mathbf{IoU}(\mathbf{:,c^{\ast}})|>1$}
	                 {\If {$\mathbf{c^{\ast}} \notin L_2$}
       	             {Number of bad predictions (Type 2) +=1\\               
       	             $L_2 = L_2 \cup \mathbf{c^{\ast}}$\\  
       	             Copy the prediction to the mask with bad predictions (Type 2)\\         	              
     	             \textbf{break}}}}}
                \ElseIf{this prediction overlaps with multiple animals}
                 {\For{each overlap with cow $\mathbf{c \in C}$}       
                   {\If {$\mathbf{c} \notin L_1$}
    	            {Number of bad predictions (Type 1) +=1\\ 
    	              $L_1 = L_1 \cup \mathbf{c}$}}   
        	    Copy the prediction to the mask with bad predictions (Type 1)}}
{\bfseries Outputs:}\\        	    
Number of good predictions, Number of bad predictions (Type 1), Number of bad predictions (Type 2)\\
Mask with good predictions, Mask with bad predictions (Type 1), Mask with bad predictions (Type 2)\\      
\caption{Pseudocode for the MaskSplitter Framework}
\label{alg:masksplitter}
\end{algorithm}
\paragraph{Overall structure of the framework}   
A unified score map is connected to the two final score maps of FCN8s with a 3x3 convolutional filter. In order to learn the three types of predictions independently, we connect three different score maps to this unified score map, one for each type, also with a 3x3 filter. We use this approach instead of a convolutional layer with 3 maps because the loss functions for each type of prediction are computed independently. To maintain spatial features and add, we take a pixelwise product of these three score maps with the split masks. The output is a sparse layer. This approach resembles a rectified linear unit, but empirically we found it to be more efficient. Each of the sparse layers is fully connected to a single unit. This architecture is designed to get the framework to learn to output the number of each type of predictions and compare it to the true number of predicted animals.    
\paragraph{Loss layers}  
These three units are compared to the true number of predicted animals using a Euclidean (L2) loss function. The fourth loss function is pixelwise sigmoid that takes the full ground truth mask and the score map for good predictions. The loss function of the full network is Equation \ref{eq:loss_masksplitter}.  
\begin{equation}
\label{eq:loss_masksplitter}
\mathcal{L} = \mathcal{L}_{gs} + \mathcal{L}_{good} + \mathcal{L}_{badcows} + \mathcal{L}_{badpreds}
\end{equation}    
here $\mathcal{L}_{gs}$ is a pixelwise cross-entropy loss taken over a sigmoid of the score map of the good predictions, the rest are Euclidean loss functions: $\mathcal{L}_{good}$ is for predicted number of good masks and true number (i.e. all animals),  $\mathcal{L}_{badcows}$ is for one mask representation overlapping with 2 or more animals and $\mathcal{L}_{badpreds}$ is for two or more mask representation overlapping with a single animal. Although losses of two types are used to compute the network's overall loss (Euclidean and Cross-Entropy), we do not normalize them by multiplying one of them by a coefficient determined using cross-validation. Although this practice is standard for many applications in deep learning literature, instance segmentation frameworks, such as \cite{he2017mask, dai2016instance}, do not use this approach.\\
\linebreak
Pseudocode for the main MaskSplitter logic is presented in Algorithm \ref{alg:masksplitter} and the flowchart in Figure \ref{fig:masksplitter}.\\ 
\section{Results}
\label{sec:results}
For experiments on benchmark datasets, we extracted 1986 images with cows from MS COCO 2017 training dataset and augmented them with 800 images of persons, sheep, horses and dogs to provide the networks with a variety of negatively labelled data. We trained the FCN8s + MaskSplitter algorithm on this combined dataset and tested them on MS COCO 2017 and Pascal VOC 2012 validation dataset.\\     
\linebreak
All models were trained with the Adam (Adaptive moment estimation) optimizer with standard parameters: $\beta_1 = 0.9, \beta_2=0.999, \varepsilon = 10^{-8}$ for 20000 iterations, base learning rate of $0.00001$ on a Tesla K40m GPU with 12 GB of VRAM. Since the network is very large ($\sim$137M parameters), we used a minibatch size 4; each iteration (feedforward + backpropagation) during training took 1.26 sec/iteration. As the testing procedure is identical to the use of FCN (with fewer score maps, 2 instead of 21), the network required only 90 ms/image (i.e. 11 images/sec) to analyze a single image. The output of our network is a binarized mask for good predictions with the same size as the input image.
Images in the MS COCO training set were cropped to 250x250 for consistency. All images in the validation set retained their dimensions. When testing on the MS COCO and Pascal VOC validation dataset, we applied the pre-trained MNC, Mask R-CNN and FCIS models as publicly available, including their configuration parameters (anchor ratios and scales, batch size, non-maximum suppression thresholds, etc).  
For comparison on benchmark datasets, we only trained our network with MaskSplitter module. Mask R-CNN, FCIS and MNC were tested out-of-the-box without any additional finetuning. This is due to the fact that these models have been extensively trained on benchmark datasets (e.g. Mask R-CNN for 160K iterations). To compare on our dataset, only Mask R-CNN, as the state-of-the-art model, was finetuned alongside our model, and these results are reported in Table \ref{tab:results_cowdata} below.
\begin{table*}
\begin{center}
\begin{tabular}{c|c|cc|cc|cc}
\hline
&Backbone Network&AP@0.5&AP@0.7&AP@0.5:0.95\\
\hline
Mask R-CNN&ResNet101&0.642&0.538&0.368\\
FCIS&ResNet101&0.633&0.527&0.327\\
MNC&VGG16&0.503&0.334&0.229\\
\hline
FCN8s+MaskSplitter&VGG16&\textbf{0.656}&\textbf{0.559}&\textbf{0.399}\\
\hline
\end{tabular}
\end{center}
\caption{Results on Pascal VOC 2012 cow validation dataset. Best results are in bold, second best italicized.}
\label{tab:pascal_val}
\end{table*}
\begin{table*}[t!]
\begin{center}
\begin{tabular}{c|c|cc|cc|cc}
\hline
&Backbone Network&AP@0.5&AP@0.7&AP@0.5:0.95\\
\hline
Mask R-CNN&ResNet101&\textit{0.549}&\textbf{0.502}&\textbf{0.341}\\
FCIS&ResNet101&\textbf{0.555}&0.314&0.229\\
MNC&VGG16&0.312&0.269&0.173\\
\hline
FCN8s+MaskSplitter&VGG16&0.443&\textit{0.321}&\textit{0.232}\\
\hline
\end{tabular}
\end{center}
\label{tab:mscoco_val_all}
\caption{Results on MS COCO 2017 cow validation dataset. Best results are in bold, second best italicized.}
\end{table*}
\begin{table*}[t!]
\begin{center}
\begin{tabular}{c|c|cc|cc|cc}
\hline
&Backbone Network&AP@0.5&AP@0.7&AP@0.5:0.95\\
\hline
Mask R-CNN&ResNet101&0.218&0.01&0.04\\
FCIS&ResNet101&0.232&0.04&0.06\\
MNC&VGG16&0.111&0.009&0.02\\
\hline
\end{tabular}
\end{center}
\caption{Results on our test dataset (before finetuning)}
\label{tab:results_cowdata_noft}
\end{table*}
\begin{table*}[t!]
\begin{center}
\begin{tabular}{c|c|cc|cc|cc}
\hline
&Backbone Network&AP@0.5&AP@0.7&AP@0.5:0.95\\
\hline
FCN8s-ft&VGG16&0.487&0.302&0.232\\
Mask R-CNN-heads-ft&ResNet101&0.637&0.253&0.265\\
Mask R-CNN-ft&ResNet101&0.687&0.259&0.298\\
\hline
FCN8s+MaskSplitter&VGG16&\textbf{0.713}&\textbf{0.505}&\textbf{0.380}\\
\hline
\end{tabular}
\end{center}
\caption{Results on our test dataset (after finetuning)}
\label{tab:results_cowdata}
\end{table*}
\subsection{Pascal VOC 2012 dataset}
Pascal VOC 2012 validation dataset contains 71 images of cows. Comparison of frameworks is presented in Table \ref{tab:pascal_val}. Although our network is conceptually simpler than Mask R-CNN, MNC or FCIS, and uses a shallow FCN8s model instead of ResNet with 101 layers, outperforms the state-of-the-art Mask R-CNN by more than 3 percentage points.
\subsection{MS COCO 2017 dataset}
MS COCO is a far more diverse dataset, in terms of the number, scale, distance and angle of the objects in images. Since both Mask R-CNN and FCIS were trained with ResNet101 as the backbone network, much of the gap between our results and theirs can be explained by ResNet's depth rather than specific strengths of Mask R-CNN and FCIS architecture. Currently fully convolutional versions of ResNet are not available, but we plan to adapt it to our architecture in the future.  
\subsection{Our Dataset}
Before finetuning Mask R-CNN to our data, we tested Mask R-CNN, FCIS and MNC on our test set. Results in Table \ref{tab:results_cowdata_noft} demonstrate that even the most advanced networks do not get good predictions due to the vast difference between the benchmark datasets and our data (see Section \ref{sec:data}). None of the networks generalized successfully to the CCTV data. Although Mask R-CNN performed slightly worse than FCIS, it showed greater capacity on other datasets, so we chose this network to finetune on our data. ResNet101-FPN (feature pyramid network) weights pre-trained on MS COCO (\url{https://github.com/matterport/Mask_RCNN/releases}, v1.0 from 23/10/2017) were selected as the baseline model, with the starting learning rate of 0.0002 and max/min image size 256x256 (as this is the closest to the 250x250 training images) for 20000 iterations. As the max/min image size was reduced, the maximal RPN anchor scale was also reduced to 256 and the number of train regions of interest (RoIs) to 32. Other configuration parameters were kept the same. 
To determine the ability of Mask R-CNN to generalize to new problems, we finetuned both the full network and only four layers that are problem-specific. Also, we finetuned the FCN8s model that does not have the capacity to separate objects. This serves as a baseline for all other models. 
Results of finetuning are presented in Table \ref{tab:results_cowdata}: FCN8s with the MaskSplitter framework confidently outperforms both Mask R-CNN models (heads-only and full): mAP is more than 8\% higher and AP@0.5 IoU is more than 2.5\% higher than the best Mask R-CNN model. This demonstrates the framework's ability to perform well on object-vs-background problems without the RPN and RoI frameworks.\\
\linebreak
The shortcoming of this part of our work was that we were able to train and test only on animals from a single feedlot, which is mainly due to the limitations of the dataset construction pipeline in \cite{ter2017bootstrapping} that fails to capture a lot of features. This is to be rectified in future publications. Another logical extension for working with video-related data that we intend to pursue is adding a temporal, i.e. a recurrent component to the network that is likely to improve the efficiency of the model on homogeneous data.         
\section{Conclusions}
\label{sec:final}
In this article we presented a conceptually simple but efficient framework that we called MaskSplitter for extracting and refining mask representation of cows, both in benchmark datasets and a challenging dataset obtained from a raw video taken at a cattle finishing facility. The framework uses both the binary mask output of the last FCN8s layer and the ground truth mask to extract information such as the number of overlaps between predicted mask representations and true objects. Without tricks like ensemble networks, image flipping and image rotation, our network performs well compared with all state-of-the-art instance segmentation solutions. This approach is easily generalizable to other backbone networks that produce image-sized score maps and other image-vs-background problems (e.g tumor detection). Other possible improvements that we intend to pursue include learning each mask prediction separately, i.e. having a map for every mask; this will allow prediction of fully separated objects and overlaps between masks.    
\section*{Acknowledgements}
Robert Ross and John Kelleher wish to acknowledge the support of the ADAPT Research Centre. The ADAPT Centre for Digital Content Technology is funded under the SFI Research Centres Programme (Grant 13/RC/2106) and is co-funded under the European Regional Development Funds. Aram Ter-Sarkisov is a recipient of Arnold G. Graves fellowship. Bernadette Earley and Michael Keane would like to thank John Horgan and Jonathan Forbes, from Kepak, for access to animals and facilities at Kepak, Co. Meath. Many thanks are due to the farm staff at TEAGASC, Grange beef research centre and at the Kepak feedlot for the care of the animals. 
\bibliography{1018}
\end{document}